\documentclass[runningheads]{llncs}

\usepackage[utf8]{inputenc}
\usepackage{amsmath}
\usepackage{booktabs}
\usepackage{multirow}
\usepackage{graphicx}

\graphicspath{{figures/}}

\begin{document}

\title{Cardiac Arrhythmia Detection from ECG with Convolutional Recurrent
    Neural Networks}
\titlerunning{Arrhythmia Detection with Neural Networks}
\author{
    Jérôme Van~Zaen\inst{1} \and
    Ricard Delgado-Gonzalo\inst{1} \and\\
    Damien Ferrario\inst{1} \and
    Mathieu Lemay\inst{1}
}
\authorrunning{J. Van Zaen et al.}
\institute{Swiss Center for Electronics and Microtechnology (CSEM),
    Neuchâtel, Switzerland}
\maketitle

\begin{abstract}
Except for a few specific types, cardiac arrhythmias are not immediately
life-threatening. However, if not treated appropriately, they can cause serious
complications. In particular, atrial fibrillation, which is characterized by
fast and irregular heart beats, increases the risk of stroke. We propose three
neural network architectures to detect abnormal rhythms from single-lead ECG
signals. These architectures combine convolutional layers to extract high-level
features pertinent for arrhythmia detection from sliding windows and recurrent
layers to aggregate these features over signals of varying durations. We
applied the neural networks to the dataset used for the challenge of Computing
in Cardiology 2017 and a dataset built by joining three databases available on
PhysioNet. Our architectures achieved an accuracy of 86.23\% on the first
dataset, similar to the winning entries of the challenge, and an accuracy of
92.02\% on the second dataset.
\keywords{Cardiac arrhythmia \and
    Machine learning \and
    Neural networks \and
    ECG.}
\end{abstract}

\section{Introduction}

Irregular electrical conduction in cardiac tissue often causes heart
arrhythmia. Atrial fibrillation is the most prevalent arrhythmia as it affects
1--2\% of the population \cite{Camm2010}. Its prevalence increases with age,
from $<$0.5\% at 40--50 years to 5--15\% at 80 years. Despite not being a
life-threatening condition from the start, it can lead to serious complications
\cite{January2014}. In particular, atrial fibrillation is associated with a
3--5 fold increased risk of stroke and a 2-fold increased risk of mortality
\cite{Kannel1998}. It was also shown to be linked with a 3-fold risk of heart
failure \cite{Wang2003}. Heart palpitations, shortness of breath, and fainting
are common symptoms. However, around one third of the cases are asymptomatic,
which prevents early diagnosis. This, in turn, delays early treatment which
might protect the patient from the consequences of atrial fibrillation and stop
its progression. Indeed, atrial fibrillation causes electrical and structural
remodeling of the atria which facilitates its further development, i.e. atrial
fibrillation begets atrial fibrillation
\cite{Wijffels1995,Frick2001,Nattel2008}.

The 12-lead ECG is the gold standard to diagnose abnormal heart rhythms. A
trained electrophysiologist can select the most appropriate therapy after
reviewing ECG signals and the patient history. This is a time-consuming task,
especially for long recordings such as the ones collected with Holter monitors.
Several approaches have been proposed to make this task easier and less
time-consuming \cite{Owis2002,DeChazal2004}. Indeed, even without perfect
accuracy, these approaches are helpful to quickly select relevant ECG segments
for in-depth analysis by a specialist.

Recently, neural networks have shown remarkable performance in numerous domains
compared to other methods. In particular, image processing was the first field
where deep neural networks surpassed existing approaches by a large margin
\cite{Krizhevsky2012}. Since then they have also been applied to multiple
signal processing classification and regression tasks with time series as
inputs. In particular, several neural networks have been proposed to detect and
classify cardiac arrhythmia from ECG signals.

In the context of the challenge of Computing in Cardiology 2017
\cite{Clifford2017}, a few approaches based on neural networks were proposed to
classify single-lead ECG signals into one of the following classes: normal
rhythm, atrial fibrillation, other rhythm, and noise. One of these approaches
applies two-dimensional convolutional layers to spectrograms computed over
sliding windows \cite{Zihlmann2017}. Aggregation of the features extracted from
the spectrograms was done either with a simple averaging over time or a
recurrent layer. However, due to convergence issues, convolutional and
recurrent layers were trained separately. A similar approach used a 16-layer
convolutional network to classify arrhythmia from ECG records
\cite{Xiong2017}. Each layer includes batch normalization, ReLU activation,
dropout, one-dimensional convolution, and global averaging.

Cardiologist-level arrhythmia detection was reached recently by a convolutional
neural network \cite{Rajpurkar2017}. This network with 34 layers was trained on
a very large dataset of 64,121 single-lead ECG signals collected from 29,163
unique patients. It can detect 12 different types of cardiac arrhythmia,
including atrial fibrillation, atrial flutter, and ventricular tachycardia.
Another approach applied convolutional neural networks to time-frequency
representations of ECG data in order to classify arrhythmia \cite{Xia2018}. Two
types of time-frequency representations were compared: the short-time Fourier
transform and the stationary wavelet transform. In this study, the second
transform led to a neural network yielding higher performance.

Thus, several neural network architectures achieved good classification
performance for the detection of abnormal heart rhythms from ECG signals.
These results are promising as they prefigure detection systems that will
quickly process long ECG records to extract pertinent segments for further
analysis by an electrophysiologist. Hopefully, this will reduce the time needed
to achieve a diagnosis and thus to select the most appropriate therapy as
early as possible. We recently proposed an approach to tackle this issue
\cite{VanZaen2019}. This approach combined a smart vest to record a single-lead
ECG over long periods of time and a convolutional recurrent neural network to
detect abnormal rhythms. In this paper, we consider variations of the neural
network architecture proposed previously and apply them to two datasets for the
classification of cardiac arrhythmia. This paper is structured as follows.
First, the datasets of ECG data and the considered neural network architectures
are presented in Section~\ref{sec:methods}. Then, the results are reported in
Section~\ref{sec:results} and discussed in Section~\ref{sec:discussion}.
Finally, a brief conclusion ends this paper in Section~\ref{sec:conclusion}.

\section{Materials and Methods}
\label{sec:methods}

\subsection{Datasets}

We trained neural networks to classify cardiac arrhythmia from ECG data with
two datasets. The first one is the dataset used for the challenge of Computing
in Cardiology 2017 \cite{Clifford2017}. It includes 8528 single-lead ECG
signals recorded with an AlivCor device. The signals are sampled at 300~Hz with
durations ranging from 9 to 60 seconds. Each record was acquired when the
subject placed their hands on the two electrodes. This resulted in a lead I
(left arm -- right arm) ECG. However, many signals are inverted (right arm --
left arm) as the device has no specific orientation.

All ECG records were labeled with one of the following four classes:
\emph{normal sinus rhythm}, \emph{atrial fibrillation}, \emph{other rhythm},
and \emph{noise}. No additional information was available about the heart
rhythms included in the \emph{other rhythm} class. The class proportions are
not balanced and vary from 3.27\% for \emph{noise} to 59.52\% for
\emph{normal sinus rhythm}. For training and evaluation, we split the
dataset into a training set with 6000 signals (70.4\%), a validation set with
1264 signals (14.8\%), and a test set with 1264 signals (14.8\%) while
approximately preserving class proportions. The full breakdown for each class
and each subset is reported in Table~\ref{table:cinc2017_breakdown}.

\begin{table}
    \centering
    \caption{Breakdown of training, validation, and test sets for the dataset
        of the challenge of Computing in Cardiology 2017.}
    \renewcommand{\arraystretch}{1.5}
    \setlength{\tabcolsep}{6pt}
    \begin{tabular}{@{}lrlrlrl@{}}
        \toprule
        Class & \multicolumn{2}{c}{Training} & \multicolumn{2}{c}{Validation}
            & \multicolumn{2}{c}{Test} \\
        \midrule
        Normal rhythm & 3571 & (59.5\%) & 752 & (59.5\%) & 753 & (59.6\%) \\
        Atrial fibrillation & 534 & (8.9\%) & 112 & (8.9\%) & 112 & (8.9\%) \\
        Other rhythm & 1699 & (28.3\%) & 358 & (28.3\%) & 358 & (28.3\%) \\
        Noise & 196 & (3.3\%) & 42 & (3.3\%) & 41 & (3.2\%) \\
        Total & 6000 & (100\%) & 1264 & (100\%) & 1264 & (100\%) \\
        \bottomrule
    \end{tabular}
    \label{table:cinc2017_breakdown}
\end{table}

The participants of the challenge of Computing in Cardiology 2017 were ranked
according to the following score evaluated on a private test set
\cite{Clifford2017}:
\begin{equation}
    S_{\text{CinC}} = \frac{F_{1n} + F_{1a} + F_{1o}}{3}
    \label{eq:cinc2017_score}
\end{equation}
where $F_{1n}$, $F_{1a}$, and $F_{1o}$ are the $F_1$ scores for
\emph{normal rhythm}, \emph{atrial fibrillation}, and \emph{other rhythm}. The
four winners \cite{Teijeiro2017,Datta2017,Zabihi2017,Hong2017} reached a score
of 0.83. It is worth mentioning that the private test set used during the
challenge was not released yet and thus could not be used for evaluation
purposes.

\begin{figure}
    \centering
    \includegraphics[width=120mm]{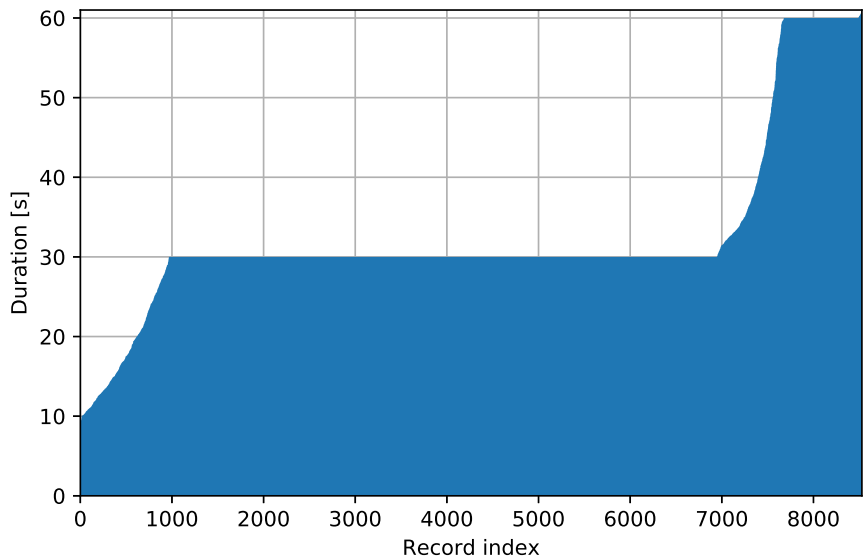}
    \caption{Durations of ECG records from the dataset of the challenge of
        Computing in Cardiology 2017 sorted in ascending order
        \cite{VanZaen2019}.}
    \label{fig:cinc2017_durations}
\end{figure}

\begin{figure}
    \centering
    \includegraphics[width=120mm]{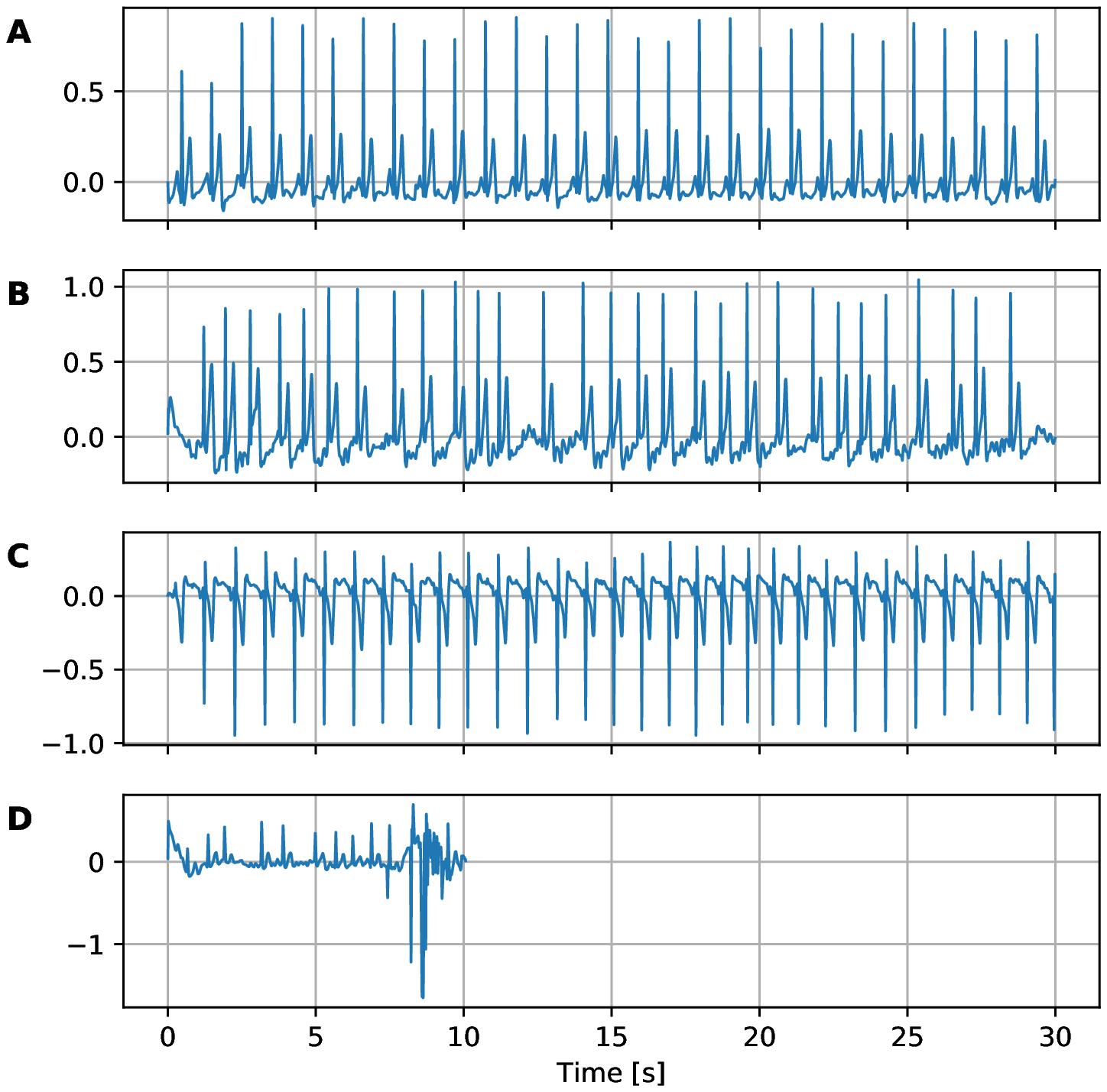}
    \caption{Examples of ECG records from the dataset of the challenge of
        Computing in Cardiology 2017: (A) normal rhythm from record A00026, (B)
        atrial fibrillation from record A00102, (C) normal rhythm from record
        A00007, (D) atrial fibrillation from record A00405.}
    \label{fig:cinc2017_examples}
\end{figure}

A number of features make this dataset challenging for cardiac arrhythmia
classification. First, as mentioned previously, many ECG signals are inverted
since the recording device lacks a clear usage orientation. Second, the
classes are not balanced. There are few records labeled
\emph{atrial fibrillation} and \emph{noise compared} to the ones labeled
\emph{normal rhythm} and \emph{other rhythm}. Third, the record durations are
not identical but instead vary between 9 and 60 seconds. These variations are
illustrated in Figure~\ref{fig:cinc2017_durations}. Most ECG signals last
around 30 seconds but a significant number have shorter or longer durations.
Furthermore, labeling is relatively coarse as there is a single label for each
ECG record. Using more than a single label would have been more appropriate as
the cardiac rhythm seems to change over the course of the several signals.
Finally, the signal quality of a non-negligible part of the records is quite
poor. Four examples are shown in Figure~\ref{fig:cinc2017_examples} to
illustrate some of these issues. The first two examples are labeled
\emph{normal rhythm} and \emph{atrial fibrillation} and their overall quality
is good. The third example is labeled \emph{normal rhythm} and has good quality
as well. However, it is inverted (R peaks are negative) compared to the first
example. In this case, the device was mostly likely held in the wrong
orientation. The last example is an example of \emph{atrial fibrillation} with
poor quality and very short duration. Indeed, the ECG signal is very noisy at
the end and seems to miss some heart beats. It also illustrates that the
records do not share the same duration. This dataset will be referred to as the
CinC 2017 dataset in the rest of the manuscript.

The second dataset we considered was built by combining three databases from
PhysioNet \cite{Goldberger2000}: the MIT-BIH Atrial Fibrillation Database
\cite{Moody1983}, the MIT-BIH Arrhythmia Database \cite{Moody2001}, and the
Long-Term Atrial Fibrillation Database \cite{Petrutiu2007}. The MIT-BIH Atrial
Fibrillation Database includes 23 two-lead ECG records sampled at 250~Hz that
last 10 hours. The MIT-BIH Arrhythmia Database is composed of 48 half-hour ECG
records with two leads collected from 47 subjects. The signals were sampled at
360~Hz. The Long-Term Atrial Fibrillation Database includes 84 two-lead ECG
records sampled at 128~Hz. These records were collected from subjects with
paroxysmal or sustained atrial fibrillation and their durations varied but were
typically between 24 and 25~hours. These three databases were annotated with
several different cardiac rhythms: atrial bigeminy, atrial fibrillation, atrial
flutter, ventricular bigeminy, heart block, idioventricular rhythm, normal
rhythm, nodal rhythm, paced rhythm, pre-excitation, sinus bradycardia,
supraventricular tachyarrhythmia, ventricular trigeminy, ventricular
fibrillation, ventricular flutter, and ventricular tachycardia.

As the ECG records from these three databases were too long to use as inputs
for neural networks, we extracted 30-second segments. Segments annotated with
more than a single label were discarded to avoid errors due to the presence of
multiple cardiac rhythms. Since the proportions of segments with
\emph{normal rhythm} and \emph{atrial fibrillation} completely dominated the
proportions for the other rhythms, we combined them in a single class labeled
as \emph{other rhythm}. Furthermore, each segment resulted in two 30-second
signals since two ECG leads were recorded in the three databases. The main
reason for keeping both leads was to test if a neural network could learn to
take into account ECG signals with different morphologies for the task of
arrhythmia detection.

The extracted 30-second ECG signals from the three databases were split into
training, validation, and test sets. We applied an iterative procedure to
assign subjects to these subsets while targeting a 60\%/20\%/20\% split and
keeping class proportions similar. This procedure was applied separately to
each database in order to approximately maintain the proportions of signals
from the three databases in the subsets for training, validation, and testing.
The rationale for this approach was to avoid any subject overlap between
the three subsets. The breakdown for each class and each
subset is summarized in Table~\ref{table:multi_breakdown_categorical}. In
addition, as the proportion of signals labeled as \emph{other rhythm} was very
low ($<$2\%), we repeated the procedure to split the data into training,
validation, and test sets while excluding this label. The objective was then
to differentiate between normal rhythm and atrial fibrillation only with a
binary classifier. In this case, the breakdown is reported in
Table~\ref{table:multi_breakdown_binary}. This dataset will be referred to as
the PhysioNet dataset from now on.

\begin{table}
    \centering
    \caption{Breakdown of training, validation, and test sets for the dataset
        combining three databases from PhysioNet with three classes.}
    \renewcommand{\arraystretch}{1.5}
    \setlength{\tabcolsep}{6pt}
    \begin{tabular}{@{}lrlrlrl@{}}
        \toprule
        Class & \multicolumn{2}{c}{Training} & \multicolumn{2}{c}{Validation}
            & \multicolumn{2}{c}{Test} \\
        \midrule
        Normal rhythm & 132474 & (44.7\%) & 43828 & (44.0\%)
            & 44080 & (44.7\%) \\
        Atrial fibrillation & 158832 & (53.6\%) & 53972 & (54.2\%)
            & 52028 & (52.8\%) \\
        Other rhythm & 4816 & (1.6\%) & 1862 & (1.9\%) & 2420 & (2.5\%) \\
        Total & 296122 & (100\%) & 99662 & (100\%) & 98528 & (100\%) \\
        \bottomrule
    \end{tabular}
    \label{table:multi_breakdown_categorical}
\end{table}

\begin{table}
    \centering
    \caption{Breakdown of training, validation, and test sets for the dataset
        combining three databases from PhysioNet with two classes.}
    \renewcommand{\arraystretch}{1.5}
    \setlength{\tabcolsep}{6pt}
    \begin{tabular}{@{}lrlrlrl@{}}
        \toprule
        Class & \multicolumn{2}{c}{Training} & \multicolumn{2}{c}{Validation}
            & \multicolumn{2}{c}{Test} \\
        \midrule
        Normal rhythm & 133150 & (45.5\%) & 43132 & (44.8\%)
            & 44100 & (45.7\%) \\
        Atrial fibrillation & 159180 & (54.5\%) & 53250 & (55.2\%)
            & 52402 & (54.3\%) \\
        Total & 292330 & (100\%) & 96382 & (100\%) & 96502 & (100\%) \\
        \bottomrule
    \end{tabular}
    \label{table:multi_breakdown_binary}
\end{table}

\subsection{Pre-processing}

After splitting both datasets into training, validation, and test sets, the
signals were pre-processed before using them as inputs to the neural networks.
The first step was to apply a digital Butterworth band-pass filter between 0.5
and 40~Hz. The filter was applied twice, once forward and once backward, to
avoid phase distortion. The specifications were chosen based on the analog
filter included in the device used to record the CinC 2017 dataset. Then, the
signals were resampled to 200~Hz in order to standardize the sampling frequency
across datasets. Finally, the signals were scaled by the mean of the standard
deviations estimated over the training set. Scaling was shown to be helpful to
accelerate training \cite{LeCun2012}. It is worth mentioning that the scaling
operation was performed separately for each database in the PhysioNet dataset
to take into account potential differences in ECG amplitude.

\subsection{Network Architectures}

Special care must be taken to handle signals with different lengths like the
ones in the first dataset. A simple solution would be to truncate all signals
to the length of the shortest one. This would make it possible to use a
convolutional network to automatically extract high-level features for
classification. However, it is not clear which part (beginning, middle, or end)
of longer signals to keep. More importantly, it would waste a huge amount of
data, especially for the first dataset where the shortest signal is around
9~seconds and the longest around 60~seconds.

A more appropriate approach is to use recurrent networks. Indeed, this class of
neural networks are well-suited to take into account sequences with different
lengths as they can, by design, remember past values for long periods of time.
However, they are not as efficient for extracting high-level features compared
to convolutional networks.

We recently proposed a neural network architecture combining convolutional and
recurrent layers to classify cardiac arrhythmia \cite{VanZaen2019}. It was
selected as it uses the strong points of both types of layers: convolutional
layers to extract high-level features and recurrent layers to handle signals
with different lengths. In this paper, we extend this architecture and test
different variations.

Each ECG signal is divided into sliding windows with 50\% overlap. We selected
two windows sizes: 512 and 1024 samples corresponding approximately to 2.5 and
5~seconds as the signals are sampled at 200~Hz. The number of windows extracted
from each signal depended on its duration. For 30-second signals, the most
common duration, this resulted in 22 windows with 512 samples and 10 windows
with 1024 samples. Convolutional layers were then applied to all windows of a
signal. Each convolutional layer is composed of a one-dimensional convolution
and a max pooling operation \cite{Zhou1988}. The convolution used a kernel of
size 5, zero padding, and a ReLU activation function \cite{Hahnloser2000}. The
max pooling operation used a pool size of 2. The first convolutional layer has
8 output channels and the subsequent layers double the number of output
channels. Therefore, the number of channels is doubled at each layer while the
window size is halved since the max pooling operation downsamples windows by
two. We tested using 7 and 8 of these convolutional layers. Then, a global
average pooling layer is applied to prepare features for the next step. The
features are fed to a long short-term memory (LSTM) layer \cite{Hochreiter1997}
with 128 units. Finally, a softmax layer outputs the probability of each class
for the input ECG windows. When training a neural network with the second
dataset without the \emph{other rhythm} class, the softmax layer is replaced by
a logistic layer since there are only two classes. The three considered
architectures are summarized in Table~\ref{table:architecture} with the
approximate numbers of parameters. It is worth mentioning that we did not try
to apply an eighth convolutional layer when using a window size of 512. The
reason is that after the seventh layer, the window size is reduced to 4. Thus,
it does not make sense to apply an additional convolutional layer with a kernel
of size 5 to such short windows.

\begin{table}
    \centering
    \caption{Neural network architectures. The output size of convolutional
        layers is given as $N \times W \times C$ where $N$ is the number of
        windows, $W$ is the window size, and $C$ is the number of channels.
        The number of classes is denoted by $K$ and the number of convolutional
        layers by $L$.}
    \renewcommand{\arraystretch}{1.5}
    \setlength{\tabcolsep}{6pt}
    \begin{tabular}{@{}lccc@{}}
        \toprule
        \multirow{2}{*}{Layer}
            & $W = 512, L = 7$
            & $W = 1024, L = 7$
            & $W = 1024, L = 8$ \\
        & Output size & Output size & Output size \\
        \midrule
        Input windows
            & $N \times 512 \times 1$
            & $N \times 1024 \times 1$
            & $N \times 1024 \times 1$ \\
        Convolutional layer 1
            & $N \times 256 \times 8$
            & $N \times 512 \times 8$
            & $N \times 512 \times 8$ \\
        Convolutional layer 2
            & $N \times 128 \times 16$
            & $N \times 256 \times 16$
            & $N \times 256 \times 16$ \\
        Convolutional layer 3
            & $N \times 64 \times 32$
            & $N \times 128 \times 32$
            & $N \times 128 \times 32$ \\
        Convolutional layer 4
            & $N \times 32 \times 64$
            & $N \times 64 \times 64$
            & $N \times 64 \times 64$ \\
        Convolutional layer 5
            & $N \times 16 \times 128$
            & $N \times 32 \times 128$
            & $N \times 32 \times 128$ \\
        Convolutional layer 6
            & $N \times 8 \times 256$
            & $N \times 16 \times 256$
            & $N \times 16 \times 256$ \\
        Convolutional layer 7
            & $N \times 4 \times 512$
            & $N \times 8 \times 512$
            & $N \times 8 \times 512$ \\
        Convolutional layer 8 & &
            & $N \times 4 \times 1024$ \\
        Global average pooling
            & $N \times 512$
            & $N \times 512$
            & $N \times 1024$ \\
        LSTM layer & 128 & 128 & 128 \\
        Softmax (or logistic) layer & $K$ (1) & $K$ (1) & $K$ (1) \\
        Number of parameters & 1.2M & 1.2M & 4.1M \\
        \bottomrule
    \end{tabular}
    \label{table:architecture}
\end{table}

\subsection{Data Augmentation}

The CinC 2017 dataset is relatively small for fitting a neural network with
only 6000 signals in the training set. Therefore, we applied two strategies to
synthetically augment the number of ECG signals available. The first strategy
is to simply flip the sign of each signal with probability 0.5. This strategy
is particularly useful for the CinC 2017 dataset where, as mentioned previously,
many signals are inverted since the recording device lacks a clear usage
orientation. Indeed, we found it easier to let the neural networks learn to
take into account inverted signals than developing an approach for detecting
and rectifying such signals before training. There is no clear justification to
apply this strategy to the PhysioNet dataset. Therefore, we trained the neural
networks for this dataset with and without random sign flipping.

The second strategy for data augmentation uses the fact that, when extracting
sliding windows, it is not possible to use all samples for the large majority
of ECG signals. Indeed, the maximum number $N$ of sliding windows of size $W$
with 50\% overlap in a signal with $M$ samples is given by
\begin{displaymath}
    N = \left\lfloor\frac{2(M - W)}{W}\right\rfloor + 1
\end{displaymath}
assuming $M \ge W$. In the previous expression, $\lfloor\cdot\rfloor$ denotes
the floor function. We took advantage of this observation to place the first
window at a random offset from the start of the signal. This random offset is
drawn uniformly from
\begin{displaymath}
    \{0, 1, 2, \dots, M - (N - 1) \cdot W / 2 - W\}
\end{displaymath}
for each signal at each epoch. The rationale behind this strategy is to prevent
the neural network from learning the precise positions of the QRS complexes in
the signals from the training set. However, to avoid wasting ECG samples, we
always used the maximum possible number of sliding windows for each signal.
Finally, it is also important to mention that these data augmentation
strategies were only applied during training and not during evaluation.

\subsection{Training}

We implemented our neural networks and the associated training pipeline with
data augmentation in Python with the Keras package \cite{Chollet2015}. We
trained the different neural network architectures for 100 epochs by minimizing
the cross-entropy with the Adam algorithm \cite{Kingma2014}. We set the initial
learning rate to 0.0005. The learning rate was divided by two if the
cross-entropy evaluated on the validation set did not decrease for 5
consecutive epochs with a lower limit at $10^{-5}$.

The batch size was set to 50 signals. We applied zero padding to ensure that
all signals in a batch had the same number of samples. Specifically, signals
that were too short were prepended with all-zero windows. To limit zero padding
as much as possible, we sorted the signals by duration and grouped them in
batches of similar lengths. This resulted in batches with varying number of
windows.

The LSTM layer was regularized by applying dropout with a rate of 0.5 to both
the input and recurrent parts \cite{Srivastava2014,Gal2016}. We monitored the
accuracy on the validation set and selected the weights at the best epoch as
the parameters for evaluation for each dataset and neural network architecture.

\section{Results}
\label{sec:results}

\begin{figure}
    \centering
    \includegraphics[width=120mm]{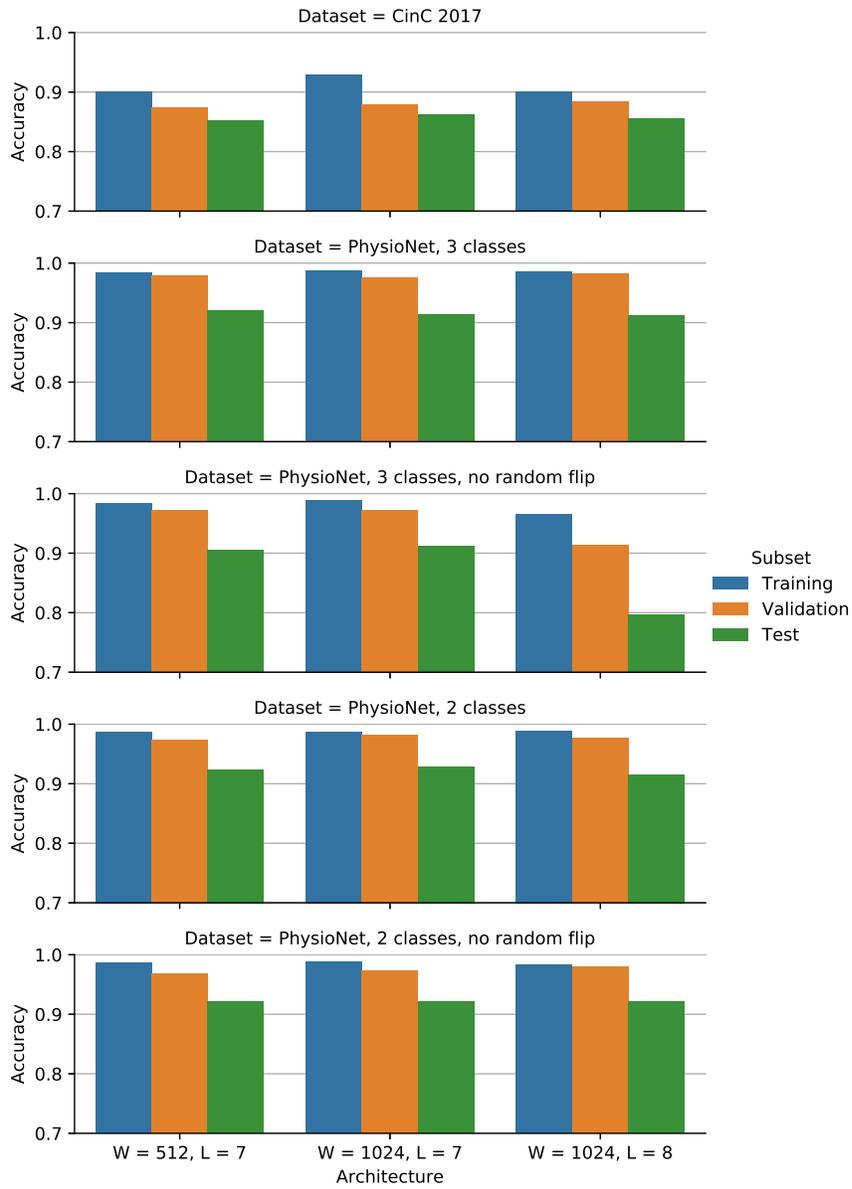}
    \caption{Cardiac rhythm classification accuracy evaluated on training,
        validation, and test sets for each dataset and each architecture. The
        window size is denoted by $W$ and the number of convolutional layers by
        $L$.}
    \label{fig:accuracy}
\end{figure}

We evaluated the three neural network architectures described in
Table~\ref{table:architecture} on the CinC 2017 and PhysioNet datasets. The
PhysioNet dataset was used with all three classes and after discarding the
\emph{other rhythm} class due its low proportion. Furthermore, we tried
training neural networks on this dataset with and without the data augmentation
strategy consisting in random flipping the sign of ECG signals. Indeed,
flipping signal signs might be detrimental to classification accuracy since the
PhysioNet dataset should not include inverted ECG records. After selecting the
best neural networks for all cases, we evaluated them without zero padding by
selecting a batch size of 1. The classification accuracy measured on the
training, validation, and test sets are shown in Figure~\ref{fig:accuracy}. In
addition, the accuracy measured on the test set is reported in
Table~\ref{table:accuracy}.

\begin{table}
    \centering
    \caption{Cardiac rhythm classification accuracy evaluated on the test
        set for each dataset and each architecture. The window size is denoted
        by $W$ and the number of convolutional layers by $L$. The best
        accuracy for each dataset is shown in bold.}
    \renewcommand{\arraystretch}{1.5}
    \setlength{\tabcolsep}{6pt}
    \begin{tabular}{@{}llccc@{}}
        \toprule
        \multicolumn{2}{@{}l}{Dataset}
            & $W = 512, L = 7$ & $W = 1024, L = 7$ & $W = 1024, L = 8$ \\
        \midrule
        \multicolumn{2}{@{}l}{CinC 2017 dataset}
            & 0.8521 & \textbf{0.8623} & 0.8560 \\
        \multicolumn{2}{@{}l}{PhysioNet dataset} \\
        & 3 classes & \textbf{0.9202} & 0.9147 & 0.9121 \\
        & 3 classes, no random flip & 0.9056 & 0.9117 & 0.7967 \\
        & 2 classes & 0.9234 & \textbf{0.9289} & 0.9149 \\
        & 2 classes, no random flip & 0.9216 & 0.9211 & 0.9221 \\
        \bottomrule
    \end{tabular}
    \label{table:accuracy}
\end{table}

The best results on the CinC 2017 dataset were obtained with a neural network
taking sliding windows with 1024 samples as input and extracting features with
7 convolutional layers. The accuracy on the test set was 86.23\%. Despite
applying dropout, there was overfitting as shown by the difference in accuracy
between the training, validation, and test sets. It also appears that using an
additional convolutional layer did not help to improve generalization
performance. By contrast, using a window size of 1024 instead of 512 was
beneficial in terms of classification accuracy. However, the performance
difference between the three considered architectures was limited to around
1\% on the test set. We also computed the score used to evaluate the
participants of the challenge of Computing in Cardiology 2017
\eqref{eq:cinc2017_score} for our best network. It achieved a score of 0.829 on
our test set which is comparable to the winning entries (0.83
\cite{Teijeiro2017,Datta2017,Zabihi2017,Hong2017}). However, it is important to
note that we could not evaluate the score on the test set used during the
challenge as it remains private at the time of writing this paper. Instead, we
had to split the official training set into smaller sets for training,
validation, and testing which reduced the available data.

We considered two cases on the PhysioNet dataset: training with three classes
(\emph{normal rhythm}, \emph{atrial fibrillation}, and \emph{other rhythm}) and
training with two classes (by discarding the class for \emph{other rhythm}). In
the first case, the best architecture used a window size of 512 and 7
convolutional layers for feature extraction and achieved an accuracy of
92.02\%. Using a larger window size or an additional convolutional layer did
not help to increase classification accuracy. In the second case, a window size
of 1024 and 7 convolutional layers led to the best performance on the test set
with an accuracy of 92.89\%. This is an expected improvement compared to the
first case since we dropped the class with the least number of signals.

A few observations can be made after reviewing the results obtained on the
PhysioNet dataset. First, it appears that randomly flipping the sign of ECG
signals during training helped to improve classification accuracy. Indeed, the
performance on the test was better for both two and three classes when this
data augmentation strategy was used during training. This result is unexpected
as the PhysioNet dataset should not include inverted ECG signals. It is
possible that this strategy, by introducing more diversity during training, led
to slightly better generalization performance.

The second observation is that there is little difference in terms of
classification accuracy between the three considered neural network
architectures. Indeed, the maximum difference was less than 2\% in all cases on
the test set. In particular, a window size of 512 was better for the case with
three classes while, in the binary case, a window size of 1024 yielded a better
classification accuracy. However, it seems that using more than 7 convolutional
layers to extract high-level features is not advantageous.

\begin{figure}
    \centering
    \includegraphics[width=120mm]{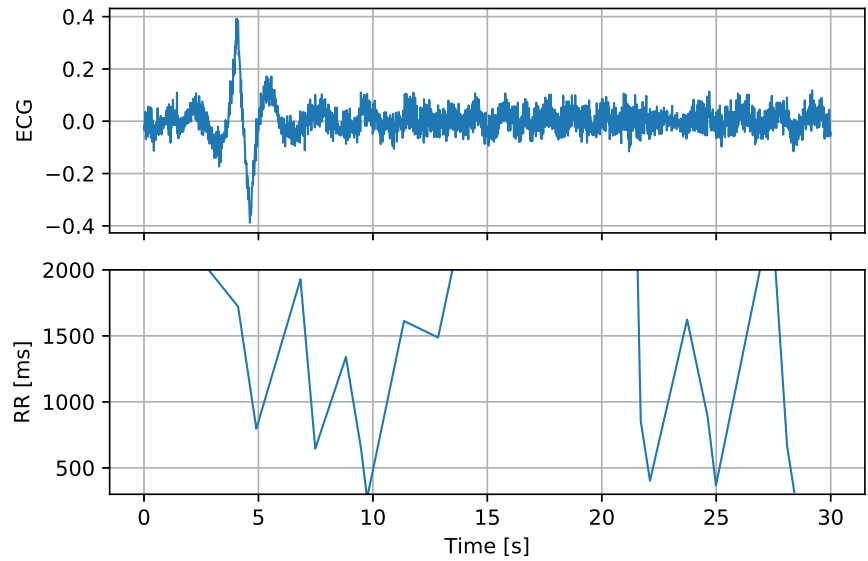}
    \caption{Example of ECG signal with poor quality labeled as
        \emph{normal rhythm} (top) and corresponding RR intervals (bottom) from
        the PhysioNet dataset. Due to poor signal quality, the RR intervals
        could not be extracted reliably.}
    \label{fig:physionet_bad_quality}
\end{figure}

\begin{figure}
    \centering
    \includegraphics[width=120mm]{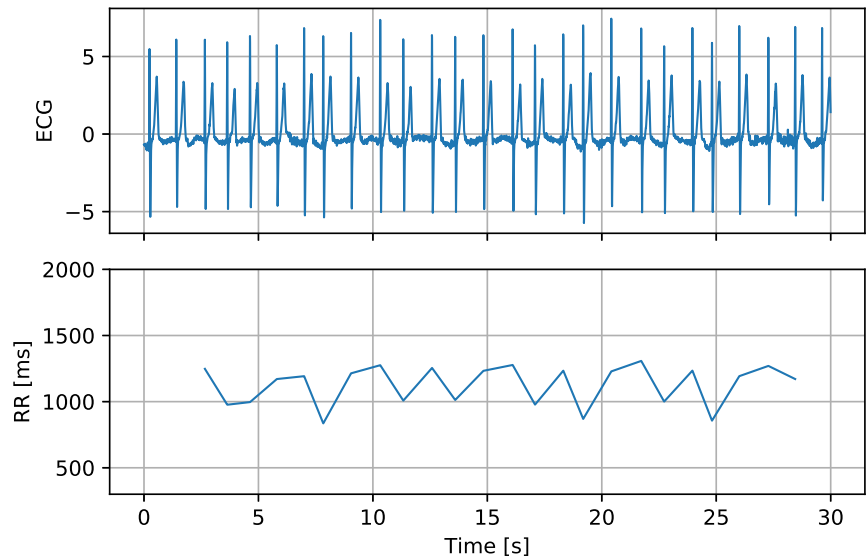}
    \caption{Example of ECG signal labeled as \emph{normal rhythm} (top) and
        corresponding RR intervals (bottom) from the PhysioNet dataset.}
    \label{fig:physionet_bad_label}
\end{figure}

The third observation that comes to mind is the large gap in accuracy due to
overfitting between training and validation sets on the one hand and test set
on the other hand. Indeed, training set accuracy was usually above 98\% and
validation set accuracy decrease only slightly while test set accuracy was 6 or
7\% lower. The small difference between the first two subsets can be explained
by the fact that we monitored performance on the validation set to select the
best weights for the neural networks. A possible explanation for the drop in
performance observed on the test set is the approach used for splitting the
original dataset. Indeed, we ensured that data for one subject was used either
for training or for evaluation (but never for both). In other words, there is
no overlap between subjects in the training, validation, and test sets. Thus,
it is possible that the ECG signals recorded from subjects assigned to the test
set are sufficiently different to cause this performance gap. It can also be
partly explained by the presence of ECG signals with poor quality in the test
set. An example of such signals is shown in
Figure~\ref{fig:physionet_bad_quality}. Due to the poor signal quality, this
signal was misclassified as \emph{atrial fibrillation} instead of
\emph{normal rhythm}. We were also unable to reliably extract the RR intervals.
Figure~\ref{fig:physionet_bad_label} shows another example of
misclassification. However, the signal quality is good in the case. It seems
the neural network predicted \emph{atrial fibrillation} instead of
\emph{normal rhythm} due to relatively due the the variations in RR intervals.
Indeed, atrial fibrillation is not associated with heart rates below 60~bpm.

Despite the observed overfitting, the classification accuracy measured on the
test set was above 90\% except for a single case (3 classes, no random sign
flipping, window size of 1024, and 8 convolutional layers). We obtained these
results on 30-second signals. A simple yet effective post-processing method to
improve classification performance would be to apply a neural network on
several consecutive 30-second segments and then pick the class with the most
predictions as the output. Of course, such an approach is only applicable when
ECG signals longer than 30~seconds are available.

\section{Discussion}
\label{sec:discussion}

The classification performance of the neural network architectures we developed
was similar to the winners of the challenge of Computing in Cardiology 2017.
However, we could only evaluate their performance on a subset of the original
training data since the official test set has not  been publicly released yet.
We also applied these network architectures to a dataset combining three
databases from PhysioNet. The classification accuracy was above 92\% when
grouping together or discarding rhythms that were neither \emph{normal rhythm}
nor \emph{atrial fibrillation}.

The three neural network architectures we considered combined convolutional and
recurrent layers. The convolutional layers were used to extract high-level
features from signal windows. Indeed, there is no need for feature engineering
with this approach as these layers learn features relevant for arrhythmia
classification during training directly from ECG data. Consequently, we applied
only a band-pass filter and scaling during pre-processing to make training
faster. The recurrent layer was used to take into account signals with
different lengths as the CinC 2017 dataset includes ECG records ranging from 9
to 60~seconds. As all signals had a duration of 30~seconds in the PhysioNet
dataset, it might have been more appropriate to avoid using recurrent layers.
However, we were interested in estimating the performance of same architectures
on a different dataset. Using only convolutional layers in this case might lead
to better performance.

We also applied two strategies for data augmentation. The first one was to
randomly flip the sign of each ECG signal during training. The main reason for
using such a strategy was to let the neural networks learn to take into account
inverted signals included in the CinC 2017 dataset. Surprisingly, this strategy
also proved to be effective for the PhysioNet dataset which should not include
inverted signals. Random sign flipping most likely helped to increase diversity
during training. The second strategy for data augmentation was to apply random
offsets from the start of each signal during training to prevent the neural
networks from learning the exact locations of QRS complexes.

Collectively, these results demonstrate that detecting cardiac arrhythmia with
neural networks from raw ECG signals is feasible. And even if classification
accuracy is imperfect, they can help to select and extract segments with
potential abnormal rhythms from long ECG recordings for further analysis by a
trained specialist. If needed, a 12-lead ECG can then be performed to confirm
or refine the diagnosis.

Despite these promising results, there is room for improvements. First, the
CinC 2017 dataset is relatively small with only 8528 records. Comparatively,
the PhysioNet dataset is much larger. However, it only includes records from
154 subjects and thus lacks diversity. In addition, several abnormal rhythms
were either grouped together or simply discarded due to the limited number of
available examples. The number of different subjects with these rhythms is even
lower. Therefore, there is a need for datasets including ECG records from a
large number of subjects with many examples of each rhythm. Obviously, this is
a difficult task as building such a dataset would be costly and time-consuming.
It is also important to note that we decided to use each lead of the PhysioNet
dataset independently in order to use the same architectures for both datasets.
Using both leads as two input channels might help to better identify abnormal
heart rhythms. In addition, as the field of neural networks is rapidly
evolving, several modifications are possible for the neural network
architectures we considered in this paper. In particular, residual connections
\cite{He2015,Xie2016} as well as dense connections \cite{Huang2017} have shown
impressive results in the context of image processing. These approaches might
also be useful for processing time series in general and ECG signals in
particular.

\section{Conclusion}
\label{sec:conclusion}

We applied three neural network architectures combining convolutional and
recurrent layers to two datasets of ECG data for the detection and
classification of cardiac arrhythmia. However, in the considered datasets,
several rhythms with only a few available examples had to be grouped together.
Future developments will need to tackle this issue by either using additional
data from other databases or by learning to recognize arrhythmia with few
examples. Furthermore, several modifications to our network architecture, such
as skip connections, might help to improve generalization performance.

\section*{Acknowledgements}

We would like to thank Clémentine Aguet and João Jorge for their helpful
comments and suggestions.

\bibliographystyle{splncs04}
\bibliography{bibliography}

\end{document}